\documentclass[10pt,twocolumn,letterpaper]{article}

\usepackage[pagenumbers]{cvpr} 

\usepackage{graphicx}
\usepackage{amsmath}
\usepackage{amssymb}
\usepackage{booktabs}
\usepackage{mathtools}
\usepackage{caption}
\usepackage{subcaption}
\DeclareMathOperator*{\argmax}{arg\,max}

\usepackage[pagebackref,breaklinks,colorlinks]{hyperref}

\usepackage[capitalize]{cleveref}
\crefname{section}{Sec.}{Secs.}
\Crefname{section}{Section}{Sections}
\Crefname{table}{Table}{Tables}
\crefname{table}{Tab.}{Tabs.}

\usepackage{cuted}
\usepackage{capt-of}
\usepackage{dsfont}

\begin{document}

\title{FairStyle: Debiasing StyleGAN2 with Style Channel Manipulations}

\author{Cemre Karakas\thanks{Equal contribution} \quad Alara Dirik\footnotemark[1] \quad Eylul Yalcinkaya\quad Pinar Yanardag\\
Boğaziçi University\\
Istanbul, Turkey\\
{\tt\small \{cemre.karakas, alara.dirik, eylul.yalcinkaya\}@boun.edu.tr, yanardag.pinar@gmail.com}
}
\maketitle

\vspace*{-\baselineskip}
\begin{strip}  \centering 
\vspace{-0.2cm}
 \begin{minipage}{.24\textwidth}    
        
        \includegraphics[width=\textwidth]{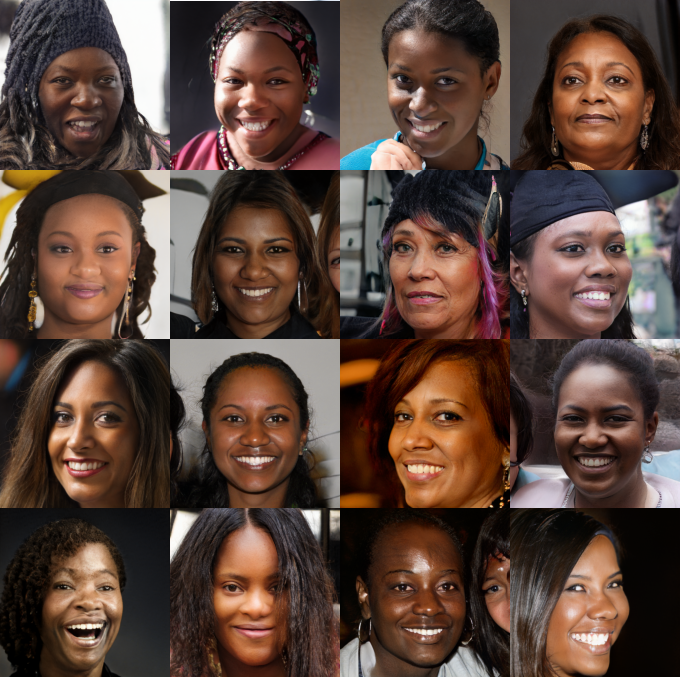}\captionof*{figure}{Black and Female}
    \end{minipage} 
    \begin{minipage}{.24\textwidth}   
          \includegraphics[width=\textwidth]{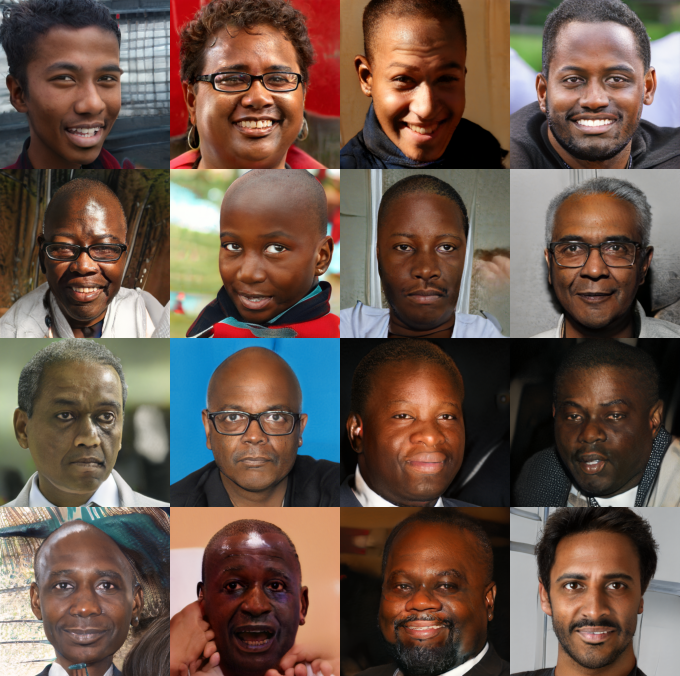} \captionof*{figure}{Black and Male}
    \end{minipage} 
    \begin{minipage}{.24\textwidth}    
    \includegraphics[width=\textwidth]{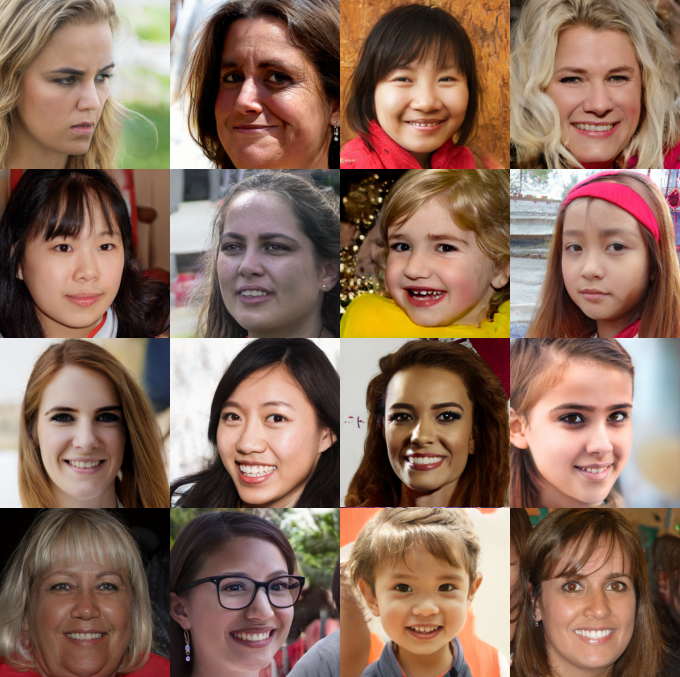}
        \captionof*{figure}{Non-Black and Female}
    \end{minipage} 
    \begin{minipage}{.24\textwidth}    
     \includegraphics[width=\textwidth]{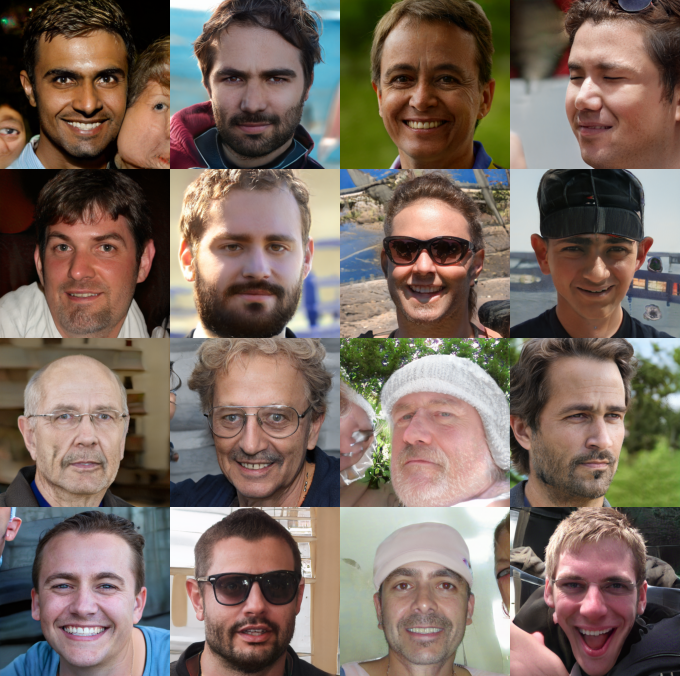}   \captionof*{figure}{Non-Black and Male}
    \end{minipage}  
    \captionof{figure}{Sample outputs from the StyleGAN2 model debiased using our method with respect to \textbf{Black+Gender} attributes.}
\label{fig:teaser}
\end{strip}

\begin{abstract}
Recent advances in generative adversarial networks have shown that it is possible to generate high-resolution and hyperrealistic images. However, the images produced by GANs are only as fair and representative as the datasets on which they are trained. In this paper, we propose a method for directly modifying a pre-trained StyleGAN2 model that can be used to generate a balanced set of images with respect to one (e.g., \textit{eyeglasses}) or more attributes (e.g., \textit{gender and eyeglasses}). Our method takes advantage of the style space of the StyleGAN2 model to perform disentangled control of the target attributes to be debiased. Our method does not require training additional models and directly debiases the GAN model, paving the way for its use in various downstream applications. Our experiments show that our method successfully debiases the GAN model within a few minutes without compromising the quality of the generated images. To promote fair generative models, we share the code and debiased models  at \small{\url{http://catlab-team.github.io/fairstyle}}.
\end{abstract}

\section{Introduction}
\label{sec:intro}
Generative Adversarial Networks (GANs) \cite{NIPS2014_5423} are popular image generation models capable of synthesizing high-quality images, and they have been used for a variety of visual applications \cite{CycleGAN, Sun_2020,DBLP:journals/corr/abs-1710-10916, wang2017highresolution, wang2019spatial,li2019single}. Like any other deep learning model, GANs are essentially statistical models trained to learn a data distribution and generate realistic data that is indistinguishable to the discriminator from that in the training set. To achieve this, GANs exploit and favor the samples that provide the most information, and may neglect minority samples. Therefore, a well-trained GAN favors learning the majority attributes, and the samples they generate suffer from the same biases in the datasets on which they are trained. For example, a GAN, trained on a face dataset with few images of non-Caucasian individuals, will generate images of mostly Caucasian individuals \cite{mcduff2019characterizing,Tan2020ImprovingTF}. Our preliminary analysis of the pre-trained StyleGAN2-FFHQ model confirms the significance of the generation bias: out of 10K randomly generated images, the \textit{male} attribute is present in 42\%, the \textit{young} attribute is present in 70\%, and the \textit{eyeglasses} attribute is present in 20\%. Our analysis shows that these biases also exist in the FFHQ training data with 42\%, 72\%, and 22\% for the \textit{male, young} and \textit{eyeglasses} attributes, respectively (see Appendix \ref{sec:appendix_prelim} for more details). These examples show that GANs not only inherit biases from the training data, but  also  carry over to the applications built on top of them. This is a particularly important issue because pre-trained large-scale GANs such as StyleGAN2 \cite{Karras2020AnalyzingAI} are often used as the backbone of various computer vision applications in a variety of domains such as image processing, image generation and manipulation, anomaly detection, dataset generation and augmentation. Therefore, any model or application that depends on large pre-trained models such as StyleGAN2 would inherit or even amplify their biases and is therefore bound to be unfair.

In this work, we aim to address the problem of fairness in GANs by debiasing a pre-trained StyleGAN2 model with respect to single or multiple attributes. After debiasing, the edited StyleGAN2 models allow the user to generate unbiased images in which the target attributes are fairly represented. Unlike previous work that requires extensive preprocessing or training an additional model for each target attribute, our approach directly debiases the GAN model to produce more balanced outputs, and it can also be used for various downstream applications. Moreover, our approach does not require any sub-sampling of the input or output data, and is able to debias the GAN model within minutes without comprimising the image quality. Our main contributions are as follows:
\begin{itemize} 
\item We first propose a simple method that debiases the GAN model with respect to a single attribute, such as \textit{gender} or \textit{eyeglasses}. 
\item We then extend our method for jointly debiasing multiple attributes such as \textit{gender and eyeglasses}.
\item To handle more complex attributes such as \textit{race}, we propose a third method based on CLIP \cite{Radford2021LearningTV}, where we debias StyleGAN2 with text-based prompts such as \textit{'a black person'} or \textit{'an asian person'}. 
\item We perform extensive comparisons between our proposed method and other approaches to enforce fairness for a variety of attributes. We empirically show that our method is very effective in de-biasing the GAN model to produce balanced datasets without compromising the quality of the generated images. 
\item To promote fair generative models and encourage further research on this topic, we provide our source code and debiased StyleGAN2 models for various attributes at \url{http://catlab-team.github.io/fairstyle}.
\end{itemize}

\section{Related Work}
\label{sec:related}
In this section, we first review related work in fairness and bias. We then discuss studies that specifically address fairness and bias in generative models. Finally, we discuss related work in the area of latent space manipulation.

\subsection {Fairness and Bias in AI}
Fairness and bias detection in deep neural networks have attracted much attention in recent years \cite{Oneto2020FairnessIM, Buolamwini2018GenderSI}. Most existing work on fairness focuses on studying the fairness of classifiers, as the predictions of these models can be directly used for discriminatory purposes or associate unjustified stereotypes with a particular class. Approaches to eliminating model bias can be divided into three main categories: Preprocessing methods that aim to collect balanced training data \cite{Liu2015DeepLF, Louizos2016TheVF, Zemel2013LearningFR}, methods to introduce constraints or regularizers into the training process \cite{Zafar2017FairnessCM, Woodworth2017LearningNP, Agarwal2018ARA}, and post-processing methods that modify the posteriors of the trained models to debias them \cite{Feldman2015ComputationalFP, Hardt2016EqualityOO}. In our work, we focus on debiasing and fairness methods developed specifically for GANs, which we discuss below.

\subsection{Detecting and Eliminating Biases in GANs}
\label{sec:gan-bias}
The fairness of generative models is much less studied compared to the fairness of discriminative models. Most research on the bias and fairness of GANs aims to either eliminate the negative effects of using imbalanced data on generation results or to identify and explain the biases. Research on bias and fairness of GANs can be divided into three main categories: improving the training and generation performance of GANs using biased datasets, identifying and explaining biases, and debiasing pre-trained GANs. 

The first research category, training GANs on biased datasets, aims to solve the problem of low quality image generation when the model is trained on imbalanced datasets with disjoint manifolds and fails to learn the true data distribution. \cite{Tanielian2020LearningDM} proposes a heuristic motivated by rejection sampling to inject \textit{disconnectedness} into GAN training to improve learning on disconnected manifolds. \cite{Tanaka2019DiscriminatorOT} proposes Discriminator Optimal Transport (DOT), a gradient ascent method driven by a Wasserstein discriminator to improve samples. \cite{Azadi2019DiscriminatorRS} uses a rejection sampling method to approximately correct errors in the distribution of the GAN generator. \cite{Grover2020FairGM} proposes a weakly supervised method to detect bias in existing datasets and assigns importance weights to samples during training. The second category of research aims to detect or explain bias in generative models. \cite{Lang2021ExplainingIS} proposes to use attribute-specific classifiers and train a generative model to specifically explain which style channels of StyleGAN2 contribute to the underlying classifier decisions. The third line of research aims to debias and improve the sample quality of pre-trained GANs. \cite{Grover2019BiasCO} proposes to train a probabilistic classifier to distinguish samples from two distributions and use this likelihood-free importance weighting method to correct for bias in generative models. However, this method requires training a classifier for each attribute targeted for debiasing and cannot handle biases in multiple attributes (e.g., \textit{gender and eyeglasses}). \cite{Tan2020ImprovingTF} proposes a conditional latent space sampling method to generate attribute-balanced images. More specifically, latent codes from StyleGAN2 are sampled and classified. Then, a Gaussian Mixture Model (GMM) is trained for each attribute to create a set of balanced latent codes. Another recent work, \cite{Ramaswamy2021FairAC}, proposes to use the latent codes from the $W$-space of StyleGAN2 to train a linear SVM model for each attribute and then use the normal vector to the separation hyperplane to steer the latent code away from or towards acquiring the target attribute for debiasing. Unlike \cite{Tan2020ImprovingTF, Ramaswamy2021FairAC}, our method does not require model training and aims to directly debias the GAN model which can be used to generate attribute-balanced image sets.

\subsection{Latent Space Manipulation}
Several methods have been proposed to exploit the latent space of GANs for image manipulation, which can be divided into two broad categories: supervised and unsupervised methods. Supervised approaches typically benefit from pre-trained attribute classifiers that guide the optimization process to discover meaningful directions in the latent space, or use labeled data to train new classifiers that directly aim to learn directions of interest \cite{goetschalckx2019ganalyze,shen2020interfacegan}. Other work shows that it is possible to find meaningful directions in latent space in an unsupervised manner \cite{voynov2020unsupervised,jahanian2019steerability}. GANSpace \cite{harkonen2020ganspace}) proposes to apply principal component analysis (PCA, \cite{wold1987principal}) to randomly select the latent vectors of the intermediate layers of the BigGAN and StyleGAN models. A similar approach is used in SeFA \cite{shen2020closed}, where they directly optimize the intermediate weight matrix of the GAN model in closed form. LatentCLR \cite{yuksel2021latentclr} proposes a contrastive learning approach to find unsupervised directions that are transferable to different classes. In addition, both StyleCLIP \cite{patashnik2021styleclip} and StyleMC \cite{Kocasari2021StyleMCMB} use CLIP to find text-based directions within StyleGAN2 and perform both coarse and fine-grained manipulations of different attributes. Another recent work, StyleFlow \cite{Abdal2021StyleFlowAE}, proposes a method for attribute-conditioned sampling and attribute-controlled editing with StyleGAN2. With respect to GAN editing, \cite{Bau2020RewritingAD} proposes a method to permanently change the parameters of a GAN to produce images in which the desired attribute (e.g., clouds, thick eyebrows) is always present. However, they did not aim to debias GANs for fairness and their methodology differs from ours.

\section{Methodology}
\begin{figure*} [t]
\begin{center}
\includegraphics[width=2\columnwidth]{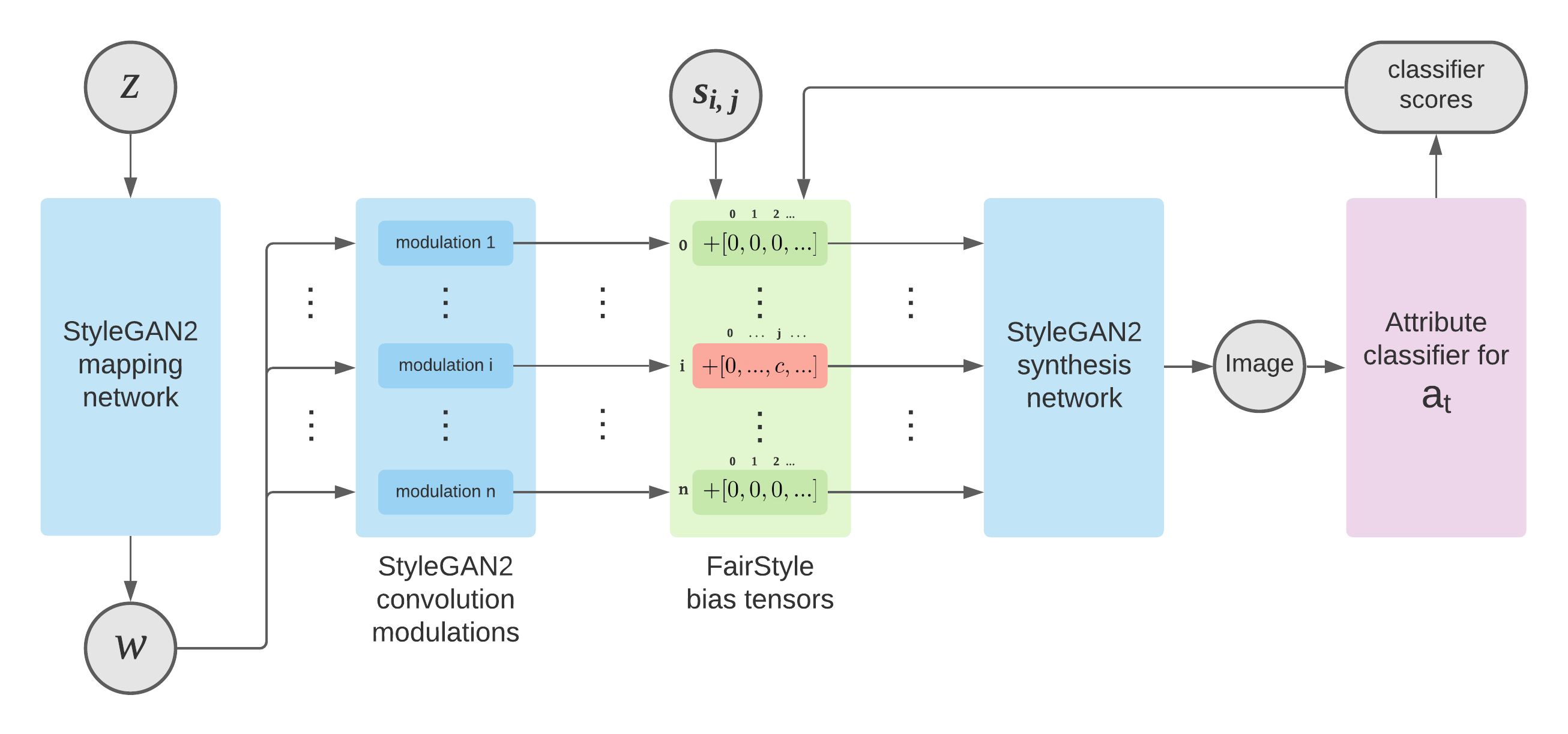}
\vskip -0.1in
\caption{An overview of the FairStyle architecture, $\mathbf{z}$ denotes a random vector drawn from a Gaussian distribution, $\mathbf{w}$ denotes the latent vector generated by the mapping network of StyleGAN2. Given a target attribute $a_t$, $s_{i,j}$ represents the style channel with layer index $i$ and channel index $j$ controlling the target attribute. We introduce \textit{fairstyle} bias tensors into the GAN model, in which we edit the corresponding style channel $s_{i,j}$ for debiasing. The edited vectors are then fed into the generator to get a new batch of images from which we obtain updated classifier results for $a_t$. The fairstyle bias tensors are iteratively edited until the GAN model produces a balanced distribution with respect to the target attribute. The de-biased GAN model can then be used for sampling purposes or directly used as a generative backbone model in downstream applications.}
\label{fig:framework}
\end{center}
\end{figure*}

In this section, we propose three methods to debias a  pre-trained StyleGAN2 model. We begin with a brief description of the StyleGAN2 architecture and then describe our methods for debiasing a single attribute, joint debiasing of multiple attributes, and debiasing with text-based  directions. Figure \ref{fig:framework} illustrates a general view of our framework. 

\subsection{Background on StyleGAN2}
\label{sec:stylegan}
The generator of StyleGAN2 contains several latent spaces: $\mathcal{Z}$, $\mathcal{W}$, $\mathcal{W+}$ and $\mathcal{S}$, also referred to as the style space. $\mathbf{z} \in \mathcal{Z}$ is a latent vector drawn from a prior distribution $p(\mathbf{z})$, typically chosen as a Gaussian. The generator $\mathcal{G}$ acts as a mapping function $\mathcal{G}: \mathcal{Z} \to \mathcal{X}$, where $\mathcal{X}$ is the target image domain. Therefore, $\mathcal{G}$ transforms the vectors from $\mathbf{z}$ into an intermediate latent space $\mathcal{W}$ by forward propagating them through 8 fully connected layers. The resulting latent vectors $\mathbf{w} \in \mathcal{W}$ are then transformed into channel-wise style parameters, forming the \textit{style space}, denoted $\mathcal{S}$. In our work, we use the style space $\mathcal{S}$ to perform manipulations, as it is shown \cite{wu2020stylespace}  to be the most disentangled, complete and informative space of StyleGAN2.

The synthesis network of the generator in StyleGAN2 consists of several blocks, each block having two convolutional layers for synthesizing feature maps. Each main block has an additional $1\times1$ convolutional layer that maps the output feature tensor to RGB colors, referred to as \textit{tRGB}. The three different style code vectors are referred to as $\mathbf{s}_{B1}$, $\mathbf{s}_{B2}$, and $\mathbf{s}_{B+tRGB}$, where $B$ indicates the block number. Given a block $B$, the style vectors $\mathbf{s}_{B1}$ and $\mathbf{s}_{B2}$ of each block consist of style channels that control disentangled visual attributes. The style vectors of each layer are obtained from the intermediate latent vectors $\mathbf{w} \in \mathcal{W}$ of the same layer by three affine transformations, $\mathbf{w}_{B1} \rightarrow \mathbf{s}_{B1}, \mathbf{w}_{B2} \rightarrow \mathbf{s}_{B2}, \mathbf{w}_{B2} \rightarrow \mathbf{s}_{B+tRGB}$.

\subsection{Measuring Generation Bias}
To assess whether our method produces a balanced distribution of attributes, we begin by formulating and quantifying the bias in the generated images. Given an $n$-dimensional image dataset $\mathcal{I} \subseteq \mathbb{R}^{n}$, GANs attempt to learn such a distribution  $P(\mathcal{I})=P_{\text {data }}(\mathcal{I})$. Thus, a well-trained generator is a mapping function $\mathcal{G}: \mathcal{Z} \rightarrow \mathcal{I}$, where $\mathcal{Z} \subseteq \mathbb{R}^{m}$ denotes the $m$-dimensional latent space, usually assumed to be a Gaussian distribution. Moreover, we can sample latent codes $\mathbf{z}$ and use the trained model to generate a realistic dataset $D=\left\{\mathcal{G}\left (\mathbf{z}_{i}\right)\right\}_{i=1}^{N}$ of $N$ generated images belonging to the distribution $P(\mathcal{I}) \approx P_{\text {data }}(\mathcal{I})$. 

Assuming that real and generated images contain $k$ semantic attributes $a_1, a_2, ..., a_k$, a well-trained GAN learns any bias inherent in the original data distribution $P_{\text {data }}(\mathcal{I})$ with respect to the semantic attributes. In our work, we are interested in finding both the marginal distribution of the individual semantic attributes $P(a_i)$ and the joint distributions of the attribute pairs $P(a_i, a_j)$ of the generated dataset $D$. To measure generation bias, we generate $N$ random images with pre-trained StyleGAN2 trained on the FFHQ dataset, and use $40$ pre-trained binary attribute classifiers \cite{StyleGAN} to assign labels to each image such that $a_i = 1$ if the image contains the attribute $a_i$, and $a_i = 0$ otherwise.

\subsection{Identifying channels that control certain attributes}
\label{sec:find-style}
For a target attribute $a_t$ such as \textit{eyeglasses}, we first propose a simple approach that identifies a single style channel $s_{i,j}$ responsible for controlling the target attribute, where layer and channel indices are denoted by $i$ and $j$, respectively. We assume that there is a binary classifier $\mathcal{C}_{a_t}$ corresponding to the target attribute, such as pre-trained CelebA binary classifiers \cite{StyleGAN}. The identified style channel $s_{i,j}$ is then used for debiasing the GAN model with respect to single (Section \ref{sec:debias-single}) and multiple attributes (Section \ref{sec:debias-multi}).
 
To identify $s_{i,j}$, we first generate $N=128$ random noise vectors to obtain their style codes using StyleGAN2. Given an arbitrary style code $\mathbf{s}$, we generate two perturbed style codes by adding and subtracting a value of $c$ at the corresponding index $i$ and channel $j$. This process is repeated for $128$ randomly generated style codes, and each perturbed style code is forward propagated through the StyleGAN2 generator to synthesize images. Finally, we identify $s_{i,j}$ corresponding to the target attribute by selecting the style channel for which the perturbation causes the highest average change in classification score over the batch of $N=128$ images:

\begin{equation}
\begin{split}
    \argmax_{i,j} \frac{\sum_{k=1}^{N} |\mathcal{C}_{a_t}(\mathcal{G}(\mathbf{s} - \Delta s_{i,j})) - \mathcal{C}_{a_t}(\mathcal{G}(\mathbf{s} + \Delta s_{i,j}))|}{N}
\end{split}
\end{equation}
 
where $\Delta s_{i,j}$ represents the perturbation value $c$, $k$ denotes the index of the generated image, and $\mathcal{G}$ denotes the generator of StyleGAN2. In other words, we repeat the same process for each channel of the style codes and leave the values of the other style channels unchanged. In our experiments, we use the perturbation value $c=10$.

\vspace{0.2em}
\noindent

\subsection{Debiasing single attributes}
\label{sec:debias-single}
Once we have identified a style channel $s_{i,j}$ that controls the target attribute $a_t$, we can perturb the value of the channel to increase or decrease the representation of the target attribute in the generated output. In our work, we use this intuition to edit the parameters of a pre-trained StyleGAN2 model that can be used to generate balanced outputs with respect to the target attribute $a_t$. 

To this end, we introduce additional bias tensors, which we call \textit{fairstyle tensors}, into the GAN model (see Figure \ref{fig:framework}). These tensors are added to the StyleGAN2 convolution modulations on a channel-wise manner. More specifically, for a fairstyle tensor, $\mathbf{b}$, we set $\mathbf{b}_{i,j}=c$ and $\mathbf{b}_{m,n} = 0$, where $m,n \neq i,j$, and $c$ is initialized to $0$. In other words, the values inside the fairstyle tensors are set to zero except for the channel indices $i,j$ that correspond to the target attribute. 

We then iteratively generate a batch of $N=128$ latent codes and compute their updated style vectors. Given an arbitrary style vector $\mathbf{s}$, we then compute the updated vector $\mathbf{s}^{\prime}=\mathbf{s}+\mathbf{b}$. We forward propagate these style vectors to generate a batch of images and compute the distribution of the target attribute using an attribute classifier. Our goal is to optimize fairstyle tensor $\mathbf{b}$ such that the images generated using the updated GAN model have a fair distribution with respect to the target attribute $a_t$. Similar to \cite{Tan2020ImprovingTF}, we use the Kullback-Leibler divergence between the class distribution of $a_t$ and a uniform distribution to compute a fairness loss value $\mathcal{L}_{\text{fair}}$, formulated as follows:

\begin{equation}
\label{eq:kl-divergence-eq}
\begin{split}
\mathcal{L}_{\text{fair}} = KL(P_{D}(a_{t})\text{ } || \text{ } \mathcal{U}(a_{t}))
\end{split}
\end{equation}

where $P_{D}$ denotes the class probability distributions and $\mathcal{U}$ denotes the uniform distribution. We used a one-dimensional gradient descent for optimizing fairstyle tensors $\mathbf{b}$. The updated GAN model with the optimized fairstyle tensors can then be used to generate images with a balanced distribution with respect to the target attribute.

\subsection{Debiasing multiple attributes}
\label{sec:debias-multi}
While our first method is effective at debiasing the GAN model with respect to a single attribute such as \textit{eyeglasses}, it does not allow for the joint debiasing of multiple attributes such as \textit{gender and eyeglasses}. Therefore, we propose to extend our method to multiple attributes. Let $a_{t_1}$ and $a_{t_2}$ represent attributes that we want to jointly debias, such as \textit{gender} and \textit{eyeglasses}. Let $s_{i_1,j_1}$ and $s_{i_2,j_2}$ represent the target style channels identified by the method in Section \ref{sec:find-style} for attributes $a_{t_1}$ and $a_{t_2}$, respectively. Similar to our first method, we iteratively generate $N=128$ random noise vectors and their corresponding style codes. Given an arbitrary style code $\mathbf{s}$, we then compute the fairstyle tensor for the corresponding channels as follows:

\begin{equation} 
\label{eq:multi-pair}
\begin{split}
\mathbf{b}_{i_1,j_1} = x_2 \times  \frac{\mathbf{s}_{i_2,j_2} - \bar{\mathbf{s}}_{i_2, j_2}}{\hat{\sigma}_{\mathbf{s}_{i_2, j_2}}} + y_2 \\
\mathbf{b}_{i_2,j_2} =   x_1 \times  \frac{\mathbf{s}_{i_1,j_1} - \bar{\mathbf{s}}_{i_1, j_1}}{\hat{\sigma}_{\mathbf{s}_{i_1,j_1}}} + y_1
\end{split}
\end{equation}

where $x_1$, $y_1$, $x_2$, $y_2$ are learned parameters initialized at $0$ and optimized using gradient descent over a batch of $N$ images, and $\bar{s}_{i,j}$, $\hat{\sigma}_{s_{i,j}}$ denote the mean and standard deviation for a given target style channel ${s}_{i,j}$ calculated as follows:

\begin{equation}
\bar{\mathbf{s}}_{i, j}=\frac{1}{N} \sum_{k=1}^{N} s_{i,j}  
\end{equation}

\begin{equation}
\hat{\sigma}_{\mathbf{s}_{i,j}}^{2}=\frac{1}{N-1} \sum_{k=1}^{N}(\mathbf{s}_{i,j}-\bar{\mathbf{s}}_{i, j})^{2}
\end{equation}
 Similar to our first method, we use KL divergence as a loss function between the joint class distribution of attributes $a_{t_1}$, $a_{t_2}$ and a uniform distribution. After optimizing the fairstyle tensor, we use the GAN model to produce a balanced distribution of images with respect to the target attributes.

Our method can also be extended to support joint debiasing for more than two attributes. Let the number of attributes for which we want to jointly debias our model be $M$ and assume that we have identified a style channel $s_{i, j}$ for each target attribute. In this case, each corresponding channel of the fairstyle tensor is updated as follows:

\begin{equation} 
 \label{eq:multi-joint}
\begin{split}
\mathbf{b}_{i_m,j_m} = \sum_{\substack{k=1, k\neq m}} ^{M}(x_{m_k} \times  \frac{\mathbf{s}_{i_k,j_k} - \bar{\mathbf{s}}_{i_k, j_k}}{\hat{\sigma}_{\mathbf{s}_{i_k, j_k}}} + y_{m_k})
\end{split}
\end{equation}

We note that Eq. \ref{eq:multi-joint} is simply a generalized version of Eq. \ref{eq:multi-pair} where each fairstyle tensor channel for a target depends on the other target channels. In this case, the number of resulting subclasses is equal to $M^2$ and the number of parameters to be learned is equal to $2 \times M \times (M-1)$.

\begin{figure*} 
\centering
    \begin{subfigure}[b]{.24\textwidth}
        \includegraphics[width=\textwidth]{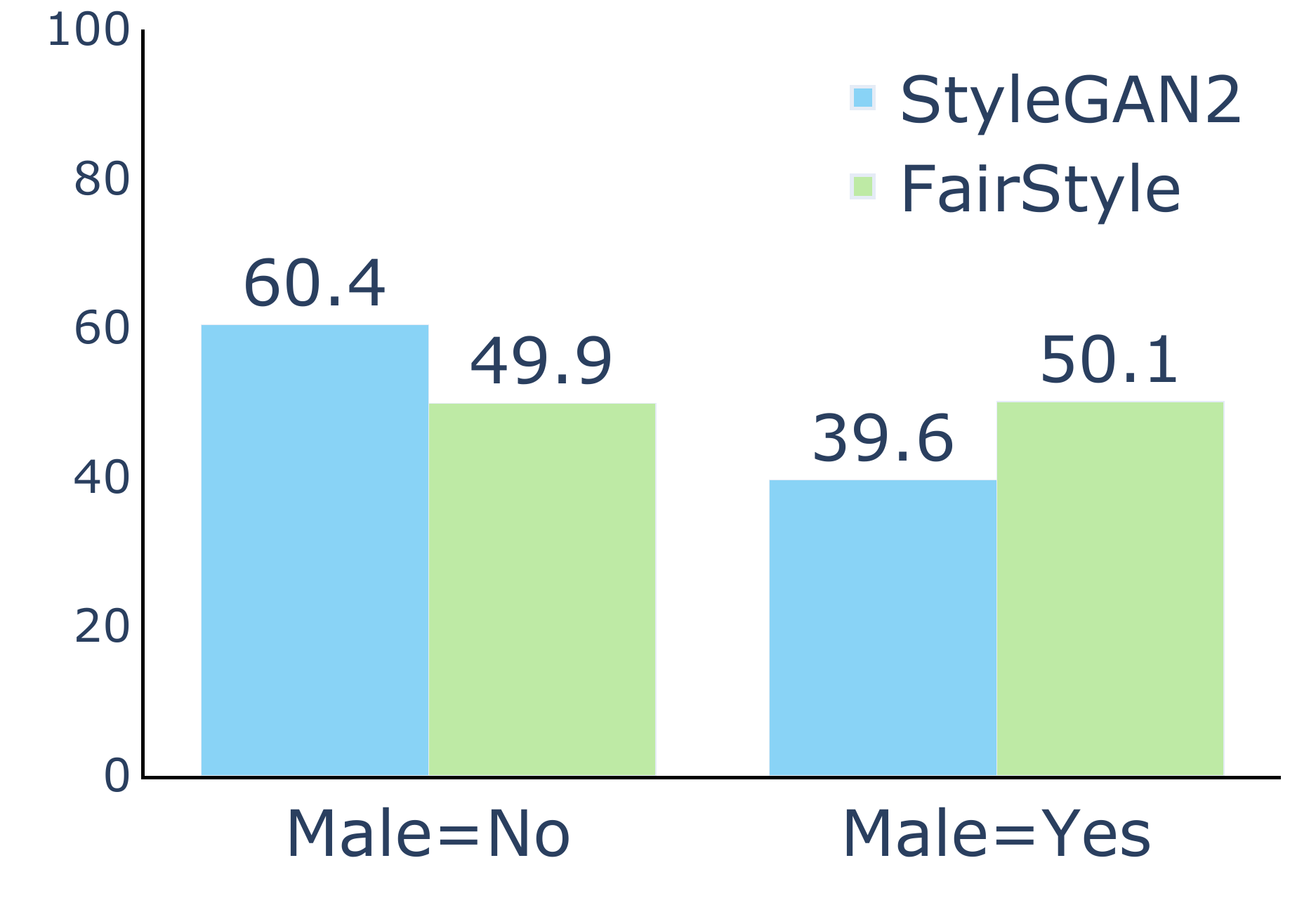}
        \caption{Gender}
    \end{subfigure}
    \begin{subfigure}[b]{.24\textwidth}    
        \includegraphics[width=\textwidth]{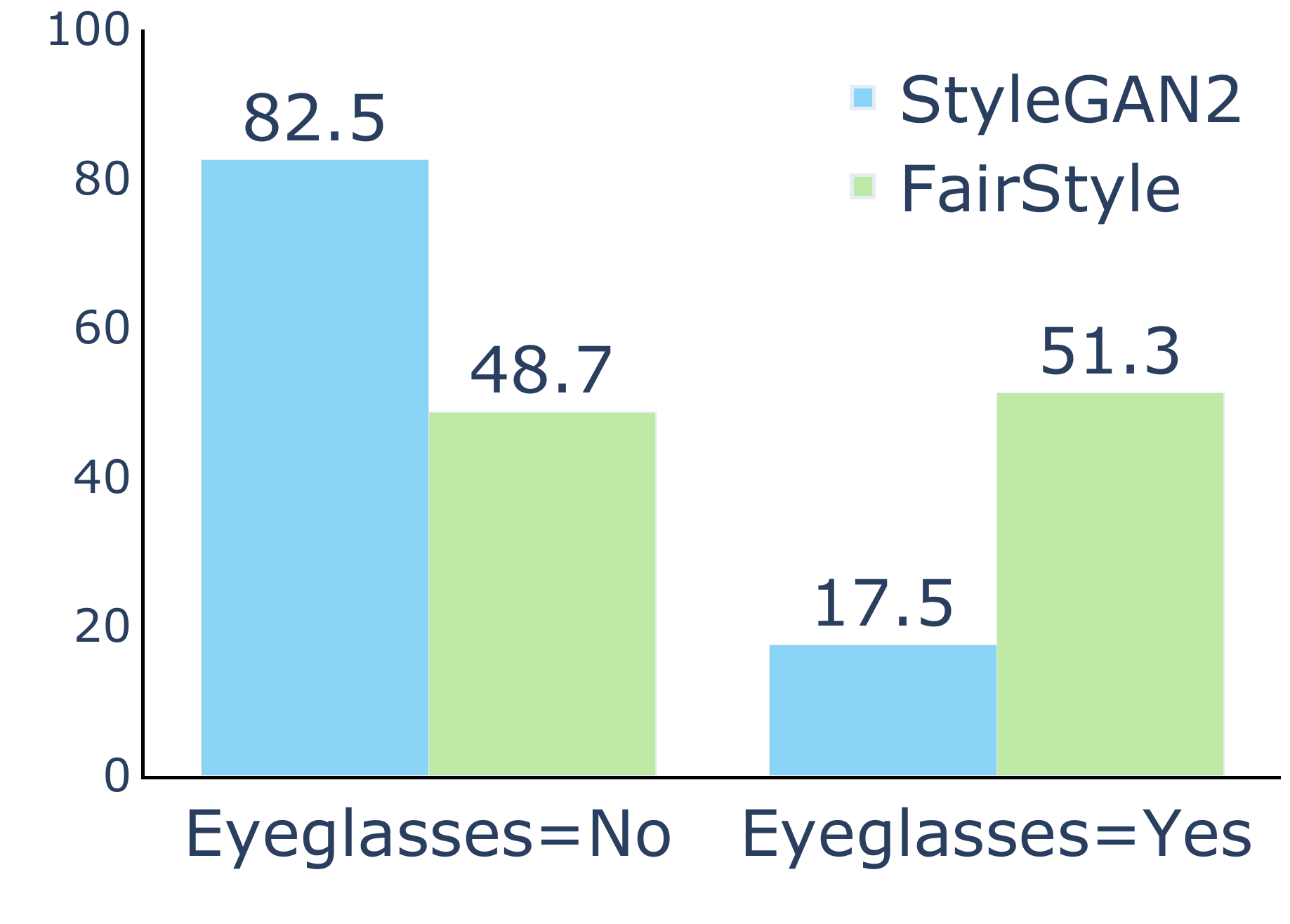}
        \caption{Eyeglasses}
    \end{subfigure} 
    \begin{subfigure}[b]{.24\textwidth}    
        \includegraphics[width=\textwidth]{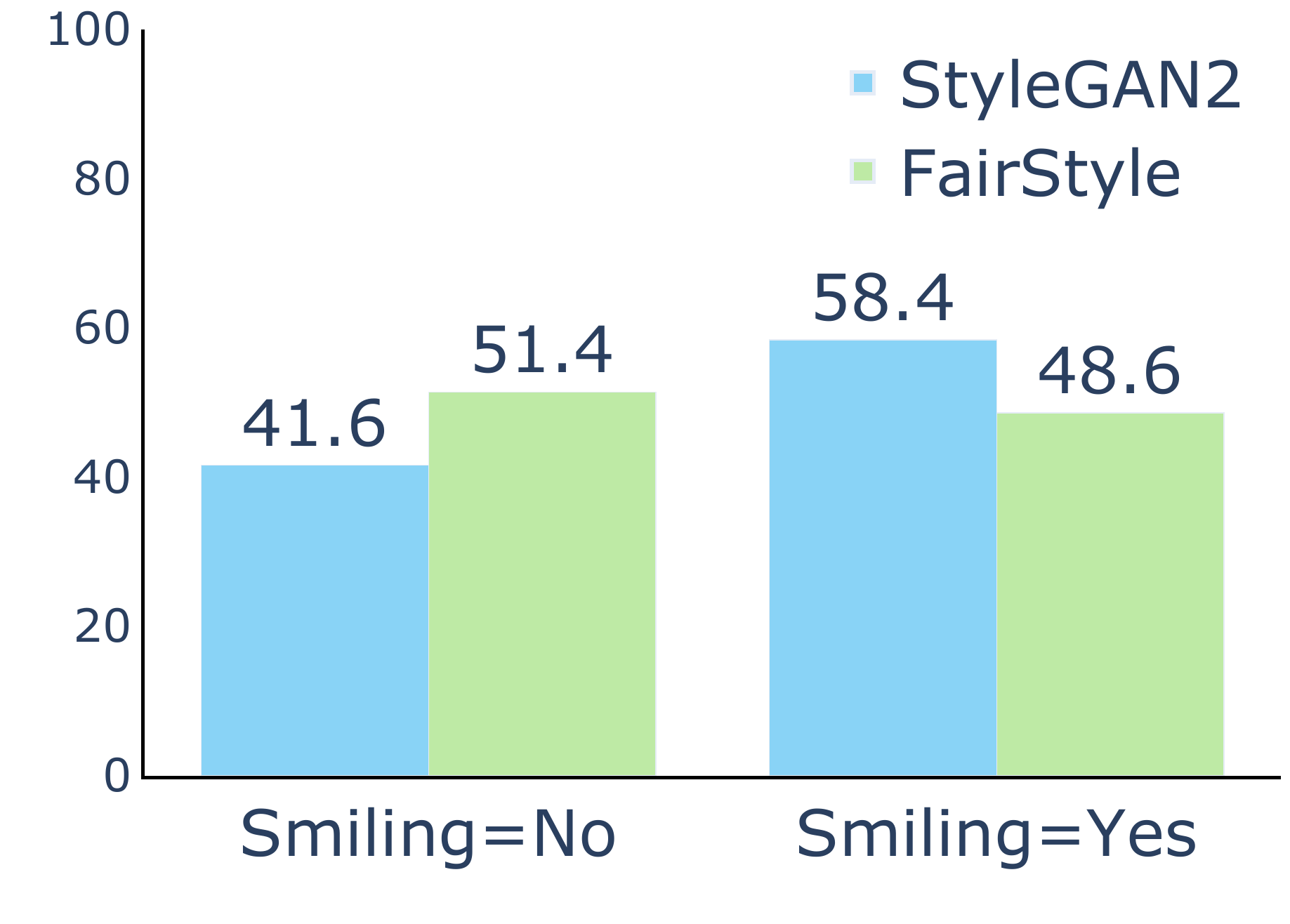}
        \caption{Smiling}
    \end{subfigure} 
    \begin{subfigure}[b]{.24\textwidth}    
        \includegraphics[width=\textwidth]{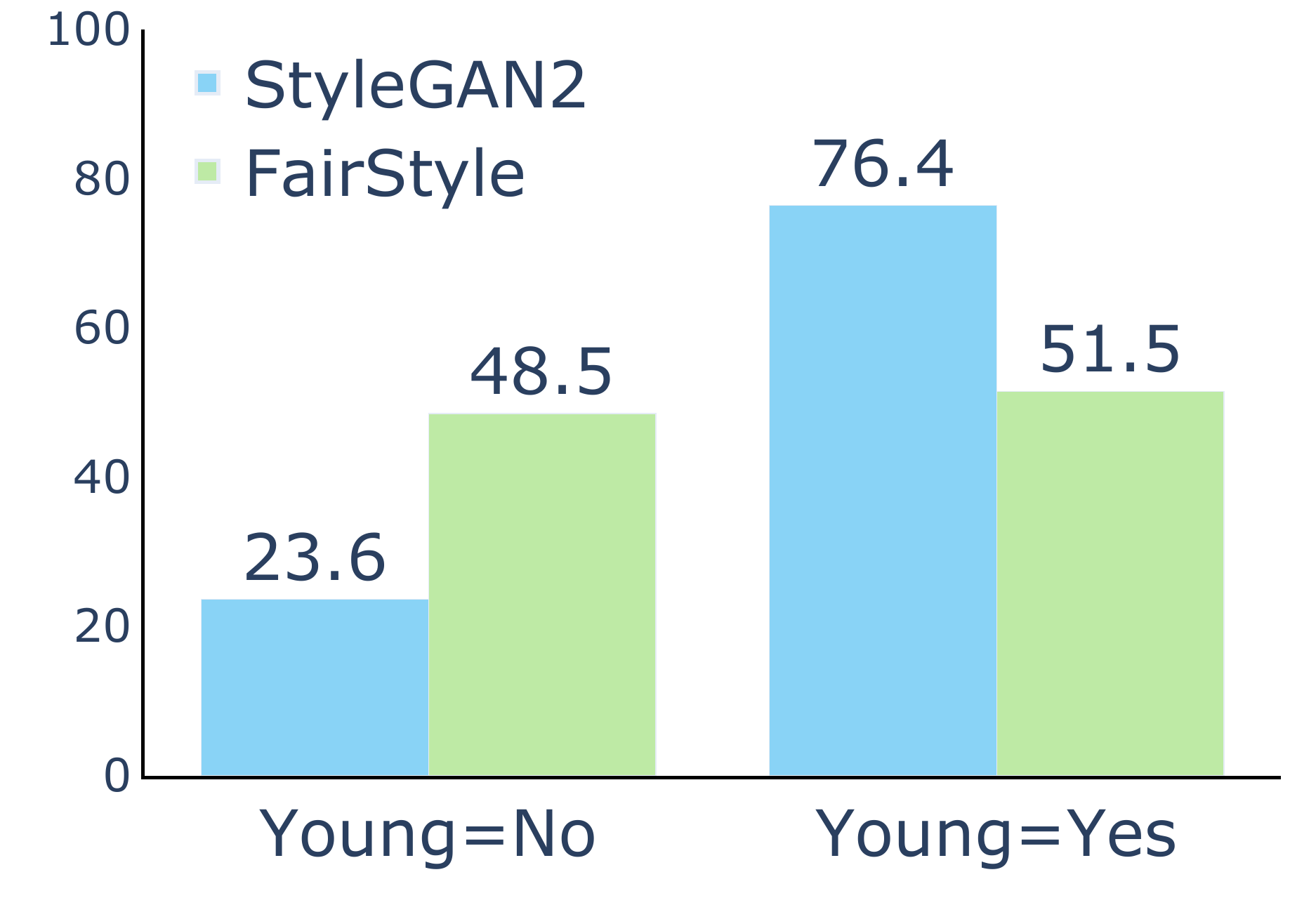}
        \caption{Young}
    \end{subfigure} 
    \begin{subfigure}[b]{.33\textwidth}    
        \includegraphics[width=\textwidth]{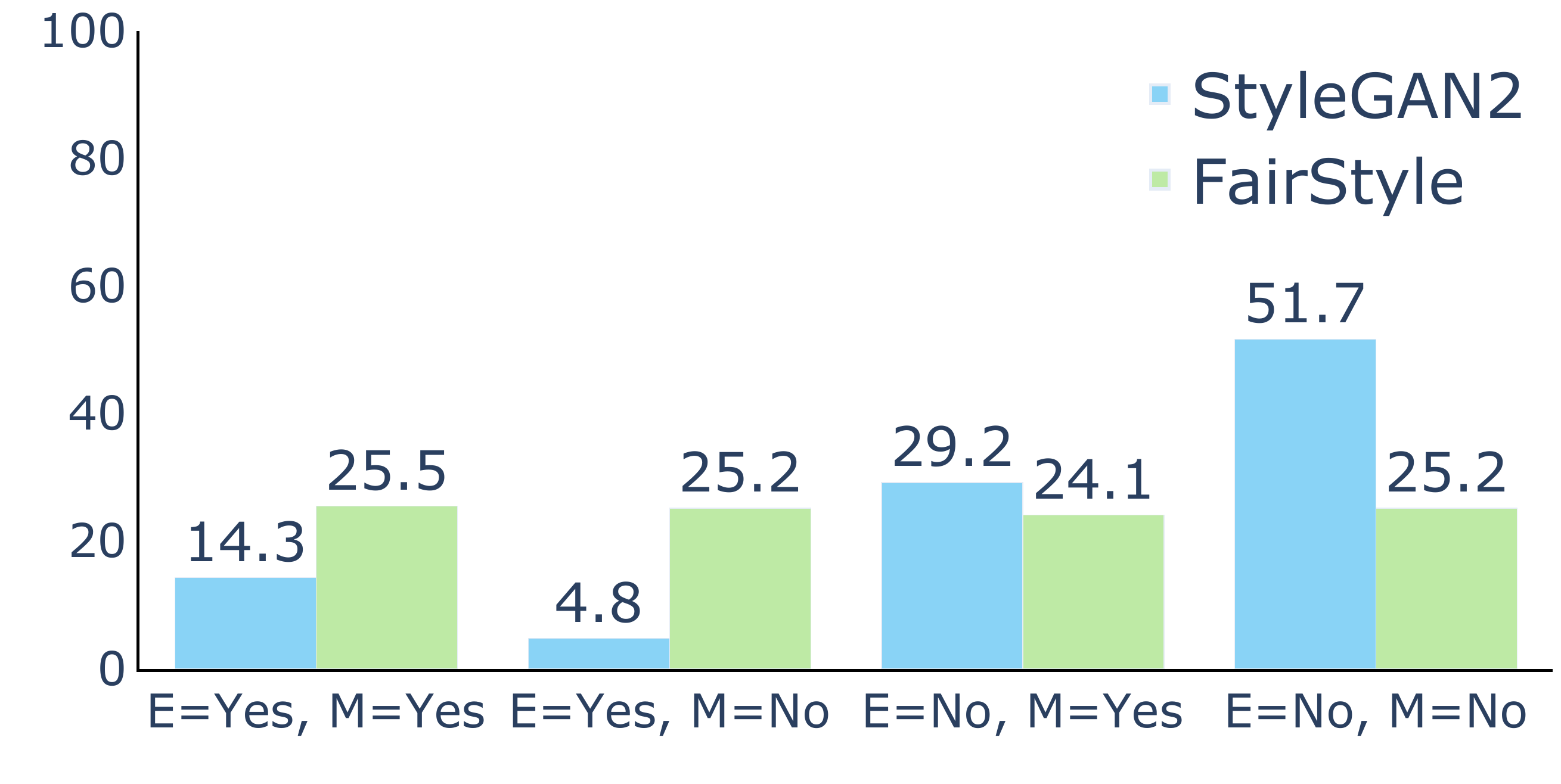}
        \caption{Gender + Eyeglasses}
    \end{subfigure} 
    \begin{subfigure}[b]{.33\textwidth}    
        \includegraphics[width=\textwidth]{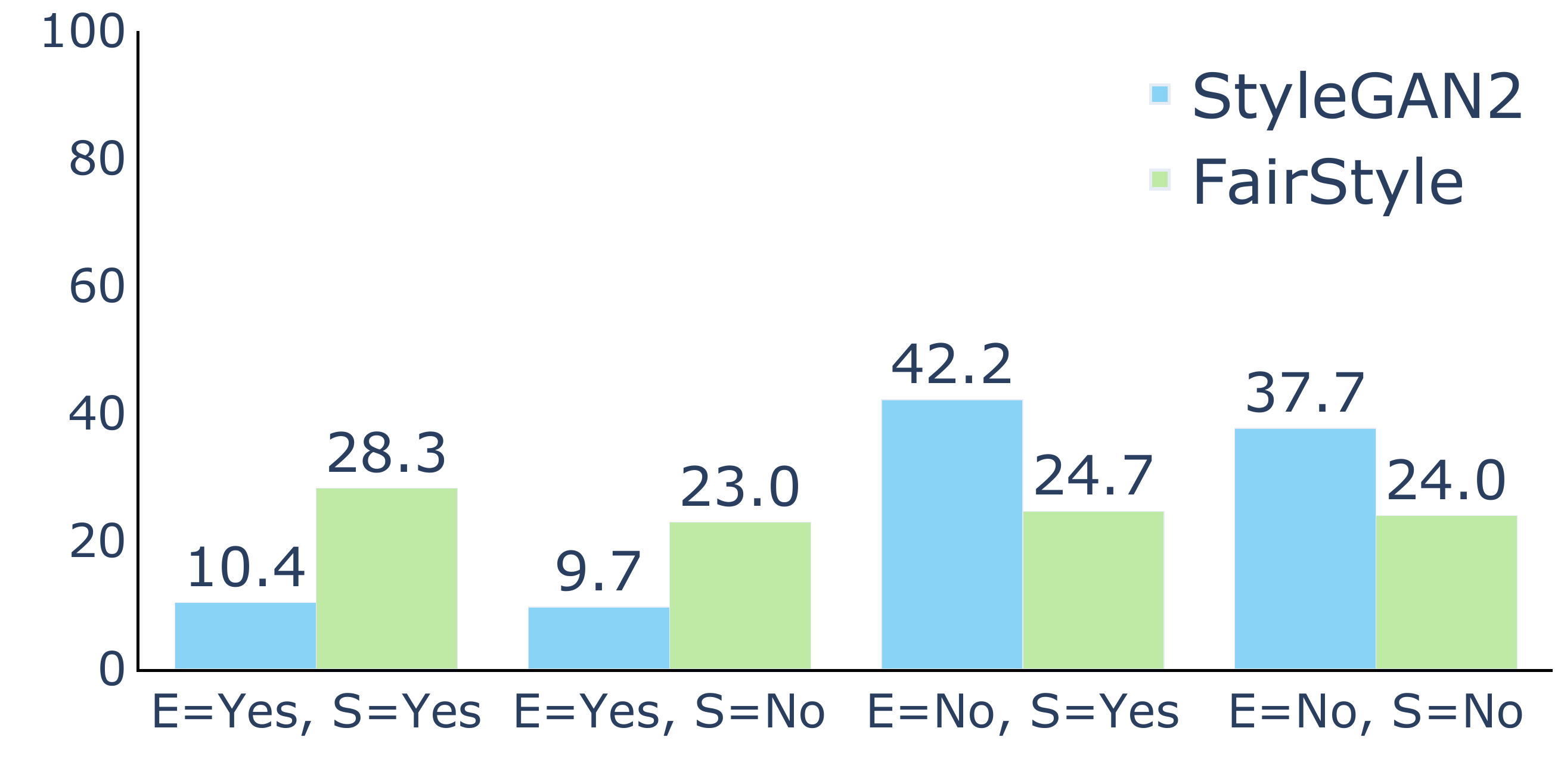}
        \caption{Eyeglasses + Smiling}
    \end{subfigure} 
    \begin{subfigure}[b]{.33\textwidth}    
        \includegraphics[width=\textwidth]{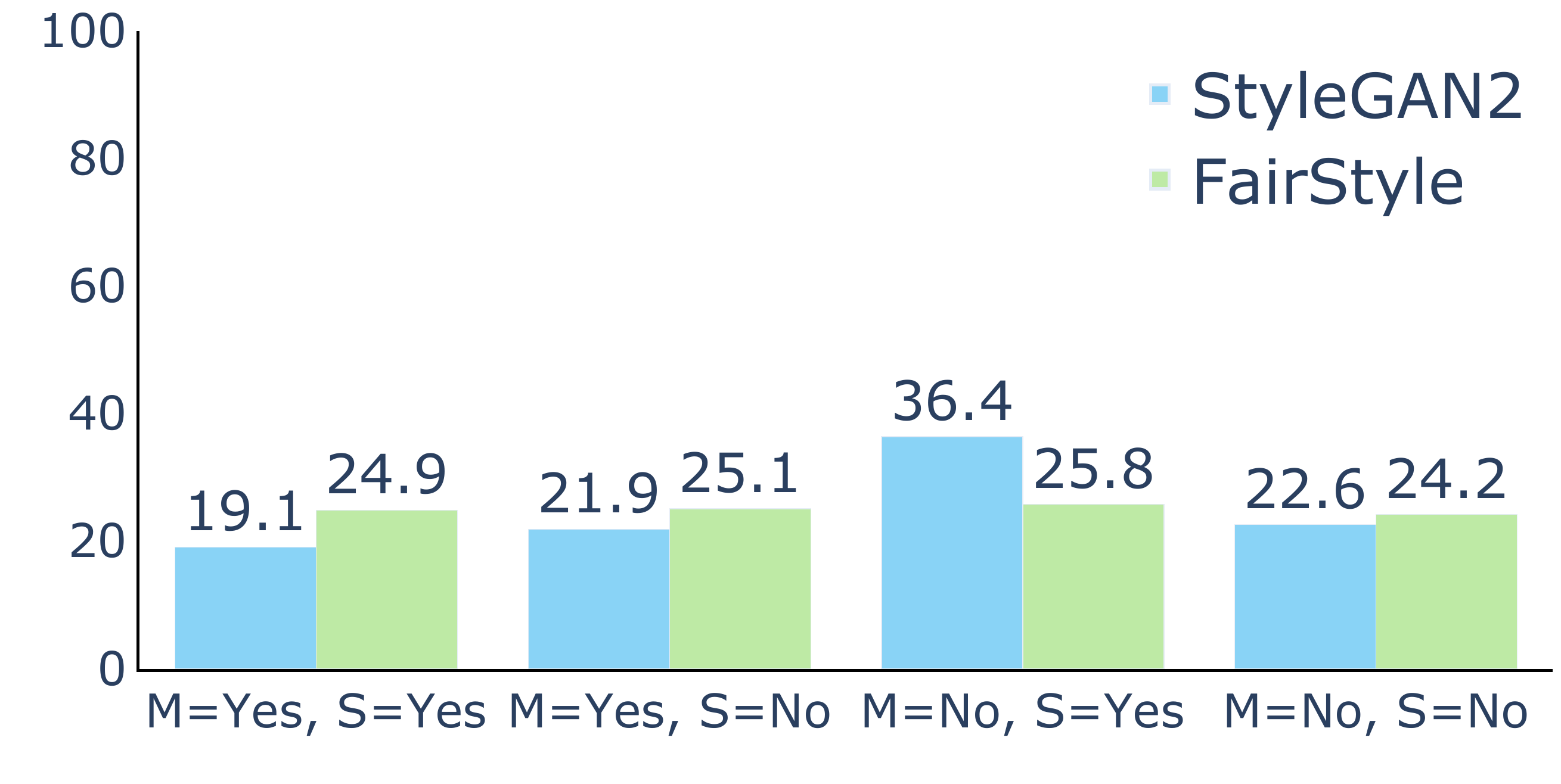}
        \caption{Gender + Smiling}
    \end{subfigure} 
    \caption{Distribution of single and joint attributes before and after debiasing StyleGAN2 model with our methods.}
    \label{fig:dist_combined}
\end{figure*}

\subsection{Debiasing attributes with text-based directions}
\label{sec:debias-clip}
The first two methods debias the GAN model with single or multiple channels, where the channels responsible for the desired attributes were identified using pre-trained attribute classifiers. However, the complexity of the attributes is limited by the availability of the classifiers. To debias even more complex attributes such as `\textit{a black person}' or `\textit{an asian person}', we debias style channels with text-based directions using CLIP. We use StyleMC \cite{Kocasari2021StyleMCMB} to  identify the individual style channels for a given text.

In addition to the text-based directions, we also replace the attribute classifier with a CLIP-based one, since binary classifiers are not available for more complex attributes. In this case, we label images by comparing their CLIP-based distances $D_{\text{ CLIP }}$ with a text prompt $a_t$ describing our target attribute and with another text prompt $a_{t_{neg}}$ negating the attribute (e.g., 'the photo of a person with curly hair' vs. 'the photo of a person with straight hair') as follows:

\begin{equation}
  \mathcal{C}_{a_t} =\begin{cases}
    1, & \text{if $D_{\text{CLIP}}(\mathcal{G}(\mathbf{s}), a_t) < D_{\text{CLIP}}(\mathcal{G}(\mathbf{s}), a_{t_{neg}}) $}.\\
    0, & \text{otherwise}.
  \end{cases}
  \label{eq:clip-label}
\end{equation}

where $\mathbf{s}$ is an arbitrary style code, $D_{\text{ CLIP }}$ is the cosine distance between CLIP embeddings of the generated image and the text prompt $a_t$ or $a_{t_{neg}}$, and $\mathcal{C}_{a_t}$ is the binary label assigned based on whichever text prompt ($a_t$ or $a_{t_{neg}}$) achieves the shortest CLIP distance from the input image. We note that the negative text prompt $a_{t_{neg}}$, as in the example above, may be biased and exclude certain groups, such as \textit{'the photo of a black person'}.

With an effective approach to assign classification scores to generated images, we identify a direction $s_{a_t}$ consisting of one or more style channels using \cite{Kocasari2021StyleMCMB}. We use the same debiasing approach as our first method by replacing $\mathbf{b}$ with $\alpha s_{a_t}$, where $\alpha$ is the hyperparameter for manipulation strength.

\begin{figure*} 
\centering
    \begin{subfigure}[b]{.24\textwidth}
        \includegraphics[width=\textwidth]{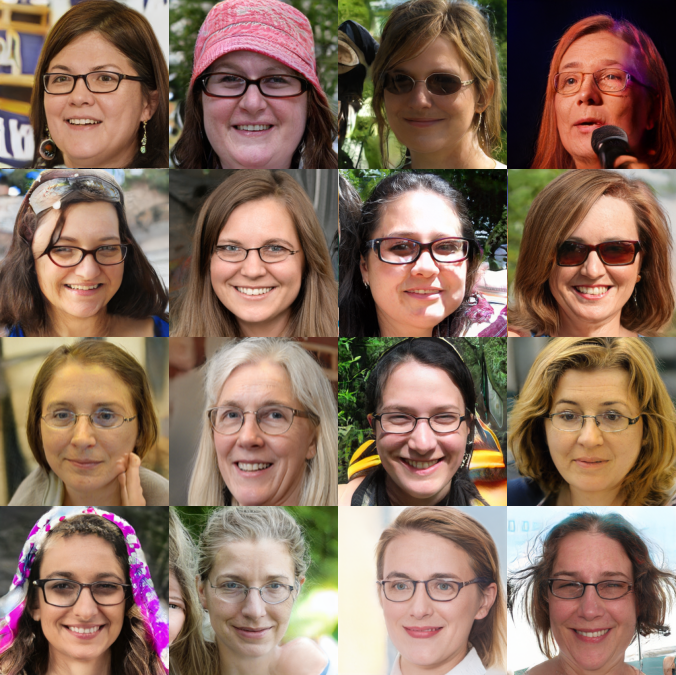}
        \caption{Female with Eyeglasses}
    \end{subfigure}
    \begin{subfigure}[b]{.24\textwidth}    
        \includegraphics[width=\textwidth]{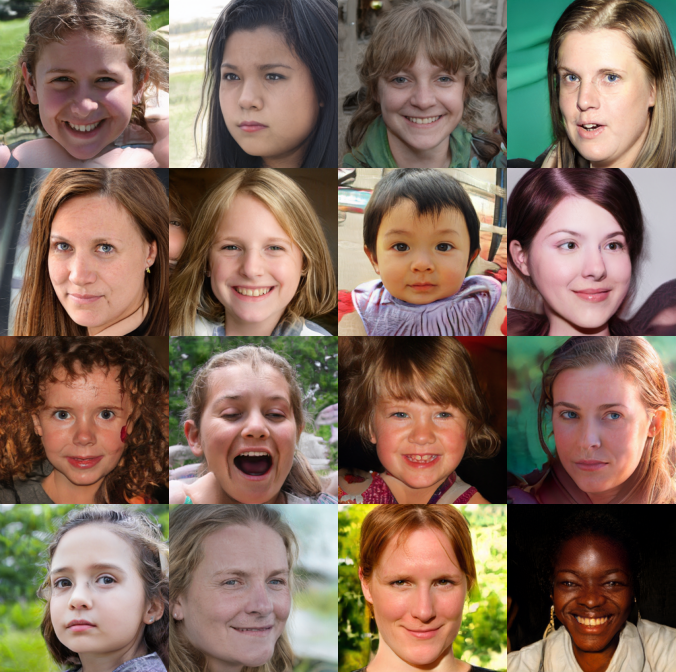}
        \caption{Female w/o Eyeglasses}
    \end{subfigure} 
    \begin{subfigure}[b]{.24\textwidth}    
        \includegraphics[width=\textwidth]{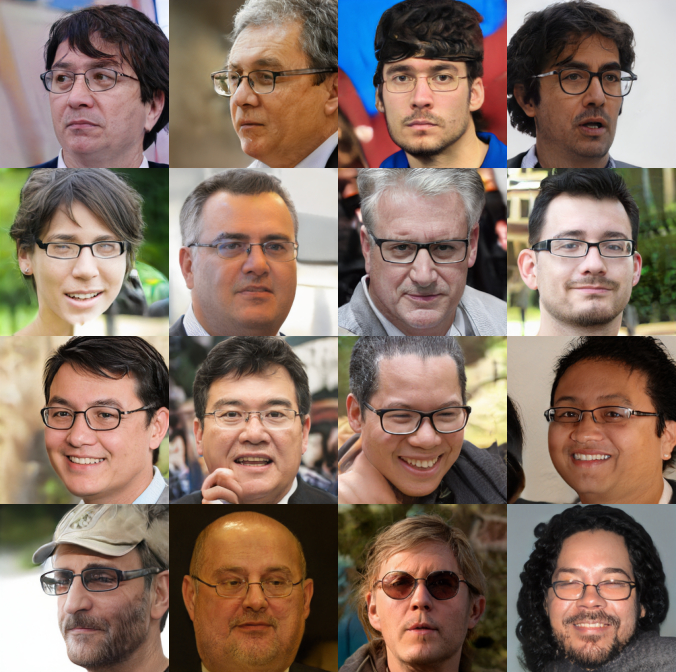}
        \caption{Male with Eyeglasses}
    \end{subfigure} 
    \begin{subfigure}[b]{.24\textwidth}    
        \includegraphics[width=\textwidth]{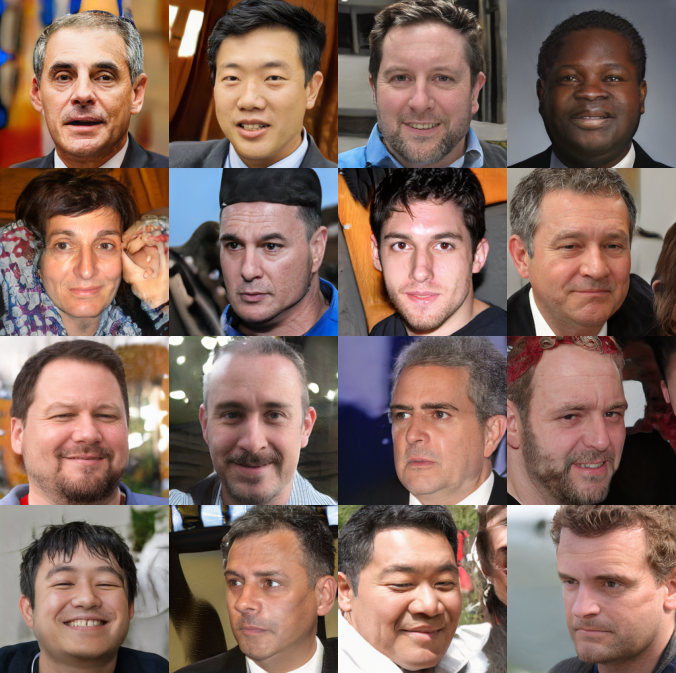}
        \caption{Male w/o Eyeglasses}
    \end{subfigure} 
    \begin{subfigure}[b]{.24\textwidth}    
        \includegraphics[width=\textwidth]{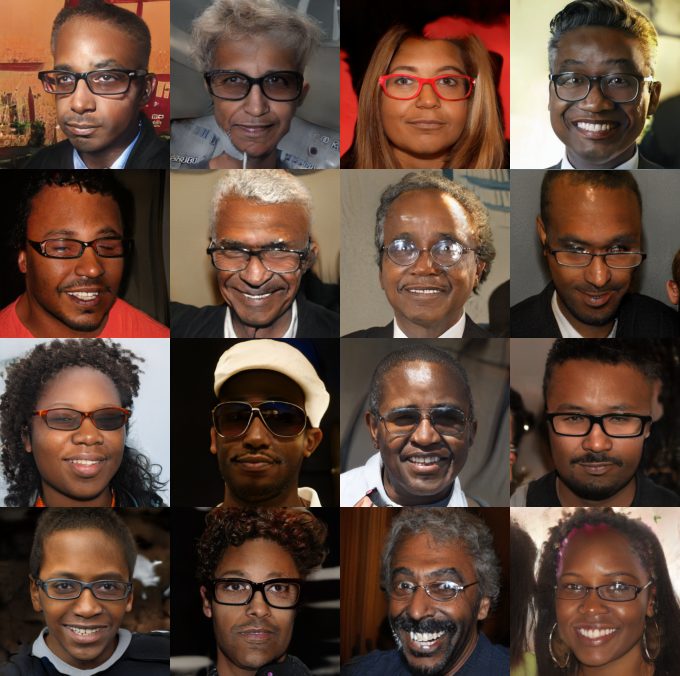}
        \caption{Black with Eyeglasses}
    \end{subfigure} 
    \begin{subfigure}[b]{.24\textwidth}    
        \includegraphics[width=\textwidth]{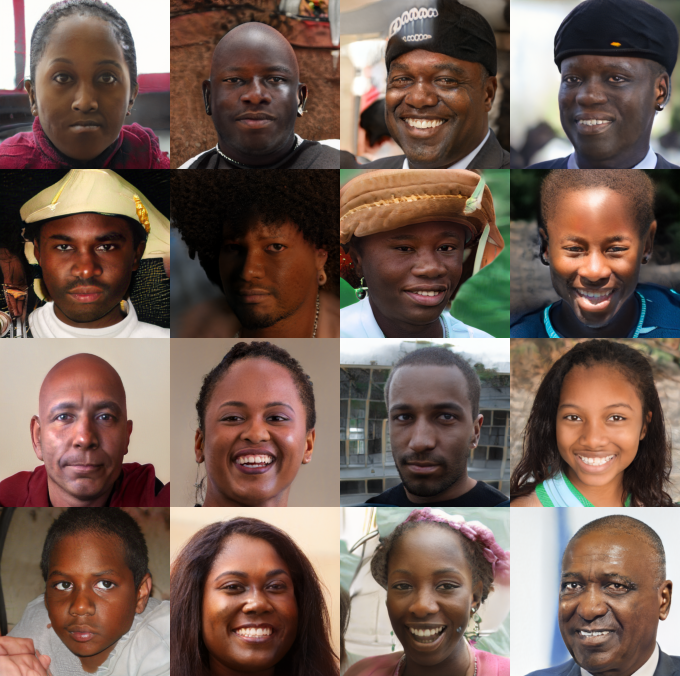}
        \caption{Black w/o Eyeglasses}
    \end{subfigure}  
    \begin{subfigure}[b]{.24\textwidth}    
        \includegraphics[width=\textwidth]{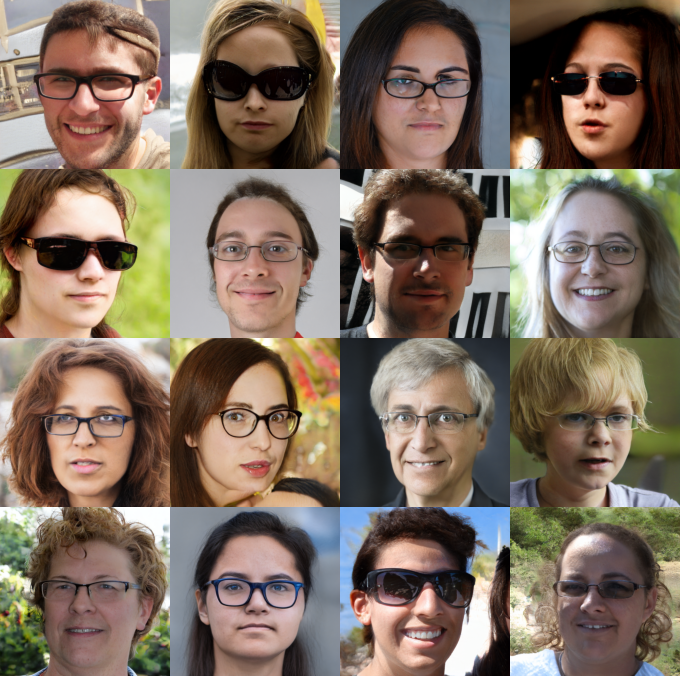}
        \caption{Non-Black with Eyeglasses}
    \end{subfigure} 
    \begin{subfigure}[b]{.24\textwidth}    
        \includegraphics[width=\textwidth]{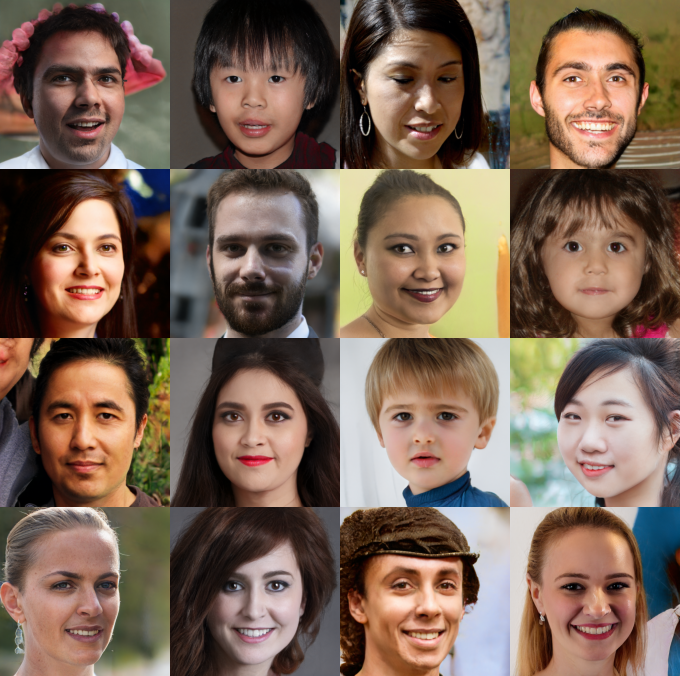}
        \caption{Non-Black w/o Eyeglasses}
    \end{subfigure}  
    \caption{Qualitative results for fair image generation in GANs with \textbf{Gender+Eyeglasses} and \textbf{Black+Eyeglasses} attributes.}
    \label{fig:qual_all}
\end{figure*}
 
\section{Experiments}
\label{sec:experiments}
In this section, we explain our experimental setup and evaluate the  proposed methods using StyleGAN2 trained on the FFHQ dataset. Furthermore, we show that our methods effectively debias  StyleGAN2 without requiring model training or affecting the quality of generation. Next, we compare our methods to FairGen \cite{Tan2020ImprovingTF} and StyleFlow \cite{Abdal2021StyleFlowAE} methods.

\subsection{Experimental Setup}
\label{sec:implementation}

For the first two methods, we identify a layer and a style channel for the \textit{gender, eyeglasses, smiling} and \textit{age} attributes and use them in our single or multiple attribute debiasing methods as described in Section \ref{sec:debias-single} and Section \ref{sec:debias-multi}. For the third method, described in Section \ref{sec:debias-clip}, we experiment with a variety of simple and complex attributes such as \textit{`a person with eyeglasses'}, \textit{`a smiling person'}, \textit{`a black person'}, \textit{`an asian person'} using \cite{Kocasari2021StyleMCMB}. We generate and label 1000 images to compute the mean and std statistics for our second method.

For our experiments, we use the official pre-trained StyleGAN2 models and binary attribute classifiers pre-trained with the CelebA-HQ dataset\footnote{\url{https://github.com/NVlabs/stylegan2}}. To identify attribute-relevant style channels, we exclude $s_{tRGB}$ layers from the style channel search since they cause entangled manipulations \cite{wu2020stylespace}. Following \cite{Kocasari2021StyleMCMB}, we also exclude the style channels of the last four blocks from the search, as they represent very fine-grained features. 

For the comparison with FairGen, we use the pre-trained GMM models\footnote{\url{https://github.com/genforce/fairgen}}. For FairGen, we had to limit our comparison to the available pre-trained models in Table \ref{tab:main_kl}. We used the StyleFlow's official implementation\footnote{\url{https://github.com/RameenAbdal/ StyleFlow}} to uniformly sample latent codes from each attribute group.  Although StyleFlow is not intended for fairness, we use it for conditional sampling similar to \cite{Tan2020ImprovingTF}.   In StyleFlow, we had to limit our comparisons to \textit{gender, smiling, eyeglasses} and \textit{age} and their multiple attributes \textit{age and eyeglasses}, \textit{age and gender}, \textit{gender and eyeglasses}. We exclude the comparison for \textit{racial attributes} for both methods because no pre-trained models were available for these attributes or training code to train new ones.

\subsection{Fairness Analysis}
To assess the fairness of the generated images, we report the KL divergence between the marginal or joint distribution of the generated images with respect to the target attributes and a uniform distribution (see Eq. \ref{eq:kl-divergence-eq}). Our goal is to obtain a  distribution with respect to one or more attributes that closely resembles a uniform distribution in order to achieve a fair distribution. To this end, we generate 10K images for each of our methods as well as for the pre-trained StyleGAN2 model, FFHQ dataset, FairGen and StyleFlow.
 
We start with our first method to debias a single target attribute, and present marginal distribution of the datasets generated with our method and the pre-trained StyleGAN2 in Figure \ref{fig:dist_combined} (a-d). As can be seen in the figure, our first method can successfully debias attributes and achieves almost perfectly balanced datasets for the attributes \textit{gender, eyeglasses, age} and \textit{smiling}. Next, we use our second method to debias \textit{gender and eyeglasses}, \textit{eyeglasses and smiling} and \textit{gender and smiling} attributes. As can be seen in Figure \ref{fig:dist_combined} (e-g), our second method is very effective at debiasing even extremely imbalanced distributions as in the case of the \textit{gender and eyeglasses} attributes, and can achieve a significant balance.

We then measure the KL divergence between the distribution of generated datasets and a uniform distribution, and provide a comprehensive comparative analysis with the FFHQ training dataset, pre-trained StyleGAN2, FairGen, and StyleFlow. We debias single attributes  for   \textit{eyeglasses}, \textit{age}, \textit{smiling}, \textit{gender} and  joint attributes for the Age+Gender, Age+Eyeglasses, and Gender+Eyeglasses (see Table \ref{tab:main_kl}). As can be seen in the table, our method outperforms StyleFlow, Fairgen and the pre-trained StyleGAN model on all attributes and achieves KL divergence values that are very close to uniform distribution in all single-attribute debiasing experiments. 

\begin{table*} [!ht]
\caption{KL Divergence between a uniform distribution and the distribution of images generated with our method, StyleFlow and FairGen. FFHQ and StyleGAN2  are included for comparison purposes.}
\vspace{0.1in}
\begin{tabular}{c|c|c|c|c|c|c|c}
\hline Method & Age+Gender & Age+Glasses & Gender+Glasses & Glasses & Age & Smiling  & Gender  \\
\hline FFHQ & 0.2456 & 0.3546 & 0.2421 & 0.186 & 0.091 & 0.005 & 0.015 \\
StyleGAN2 & 0.2794 & 0.3836 & 0.2495 & 0.180 & 0.109 & 0.011 & 0.018\\
\hline 
StyleFlow & 0.2141 & 0.1620 & 0.1214 & 0.061 & $3.98\times10^{-4}$ & 0.045 & 0.023 \\
FairGen & $3.73\times10^{-2}$ & $3.30\times10^{-2}$ & $1.85\times10^{-3}$ & $7.07\times10^{-4}$ & $1.77\times10^{-3}$ & $1.80\times10^{-5}$ & $4.21\times10^{-4}$\\
{\bf FairStyle} & $2.57\times10^{-2}$ &  $1.57\times10^{-2}$ & $2.41\times10^{-4}$ & 0 & $1.80\times10^{-7}$ & $8\times10^{-8}$ & $3.20\times10^{-7}$ \\
\hline
\end{tabular}
\label{tab:main_kl}
\end{table*}
We also perform additional single-attribute debiasing experiments for the highly biased attributes \textit{black}, \textit{asian}, and \textit{white}. Since the CelebA classifiers did not cover these attributes, we used our CLIP-based method to debias the StyleGAN2 model for the \textit{black}, \textit{asian}, and \textit{white} attributes. We present the results of this experiment in Table \ref{tab:kl-div-single-race}. As can be seen in the table, our method achieves a distribution that is very close to a uniform distribution, and effectively produces unbiased datasets with respect to the racial attributes.\\

\subsection{Qualitative Results}
 We use our methods to debias StyleGAN2 for multiple attributes and show the generated images in Figure \ref{fig:teaser} and Figure \ref{fig:qual_all}. As can be seen in the figures, our method is able to generate balanced images for the attributes \textit{gender with eyeglasses} (Figure \ref{fig:qual_all} (a-d)), \textit{gender and black} (Figure \ref{fig:teaser} (a-d)) and attributes \textit{black and eyeglasses} (Figure \ref{fig:qual_all}) (e-h)).\\
 
\subsection{Runtime Analysis}
Our method directly debias the StyleGAN2 model  within a short period of time. More specifically, the average time to debias a single attribute is 2.25 minutes, while debiasing joint attributes takes 4.2 minutes. \\

\subsection {Generation Quality}
We note that a fair generative model should not compromise on generation quality to maintain its usefulness. To ensure that our methods generate high quality and diverse images, we  report the Fréchet Inception Distance (FID) between sets of $10K$ images generated by the debiased StyleGAN2 model produced by our method and by the pre-trained StyleGAN2 model. Unlike our method, FairGen and StyleFlow do not edit the GAN model, but rely on subsampling latent vectors from GMM or normalizing flows models. Therefore, we exclude them from the FID experiments. 

To test image quality after debiasing the GAN model, we use the attribute pairs \textit{gender and eyeglasses}, \textit{race and gender} and \textit{race and eyeglasses} to compute the FID scores of the debiased datasets. While the pre-trained StyleGAN2 model achieves a FID score of $14.11$, our method achieves fairly similar FID score of $14.72$ (a lower FID score is better). Note that a small increase in FID scores is expected as the distribution of generated images is shifted for debiasing compared to the real images from the training data. However, we note that the increase in FID score is negligible and the debiased GAN model still generates high quality images (see Figure \ref{fig:teaser} and Figure \ref{fig:qual_all}).\\

\begin{table}
\caption{KL Divergence between a uniform distribution and the distribution of images generated by our text-based method to debias the \textit{black}, \textit{asian}, and \textit{white} attributes. FFHQ and StyleGAN2 are included for comparison purposes.}
\vspace{0.1in}
\begin{tabular}{c|c|c|c}
\hline Method & Black & Asian & White  \\
\hline FFHQ & 0.576 & 0.279 & 0.042 \\
StyleGAN2 & 0.603 & 0.319 & 0.057 \\
\hline
\bf FairStyle & $8.00 \times 10^{-6}$ & $7.20 \times 10^{-7}$ & $2 \times 10^{-6}$ \\
\hline
\end{tabular}
\label{tab:kl-div-single-race}
\end{table}

\section{Limitations and Broader Impact}
\label{sec:limitations}
While our proposed method is effective in debiasing GAN models, it requires pre-trained attribute classifiers for style code optimization. We note that the debiasing process can be affected by biases in these classifiers, a problem that also occurs in the competing methods. This is especially important when debiasing attributes that are known to be biased, such as racial attributes like \textit{black} or \textit{asian}.\\

\section{Conclusion}
\label{sec:conclusion}
Generative models are only as fair as the data sets on which they are trained. In this work, we attempt to address this problem and propose three novel methods for debiasing a pre-trained StyleGAN2 model to allow fairer data generation with respect to a single or multiple target attributes. Unlike previous work that requires training a separate model for each target attribute or subsampling from the latent space to generate debiased datasets, our method restricts the debiasing process to the style space of StyleGAN2 and directly  edits the GAN model for fast and stable fair data generation. In our experiments, we have shown that our method is not only effective in debiasing, but also does not affect the generation quality. 

We believe that our method is not only useful for generating fairer data, but also  our debiased models can serve as a fairer framework for various applications built on StyleGAN2. We hope that our work will not only raise awareness of the importance of fairness in generative models, but also serve as a foundation for future research.\\

\noindent \textbf{Acknowledgments}
This publication has been produced benefiting from the 2232 International Fellowship for Outstanding Researchers Program of TUBITAK (Project No: 118c321). We also acknowledge the support of NVIDIA Corporation through the donation of the TITAN X GPU and GCP research credits from Google.  

\newpage

{\small
\bibliographystyle{ieee_fullname}
\bibliography{egbib}
}

\appendix

\section{Fairness Analysis on FFHQ Data and StyleGAN2 FFHQ Model} \label{sec:appendix_prelim}
\begin{figure*}
\begin{center}
\centerline{
\includegraphics[width=.25\textwidth]{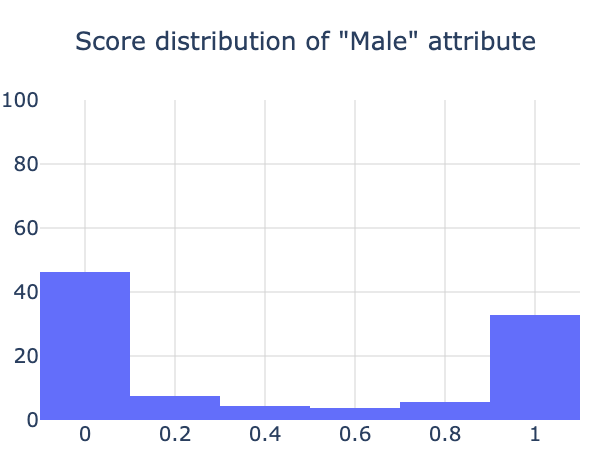}
\includegraphics[width=.25\textwidth]{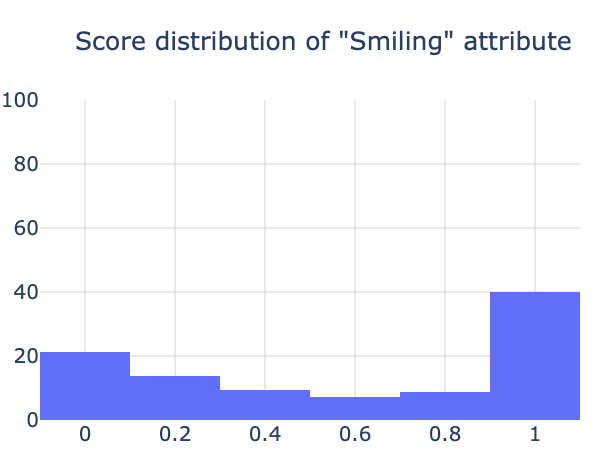}
\includegraphics[width=.25\textwidth]{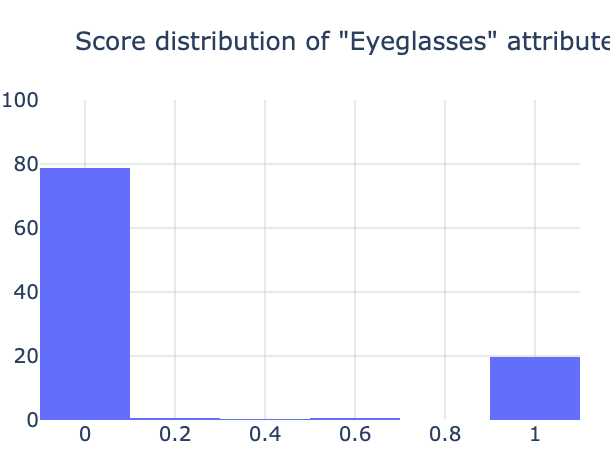}
\includegraphics[width=.25\textwidth]{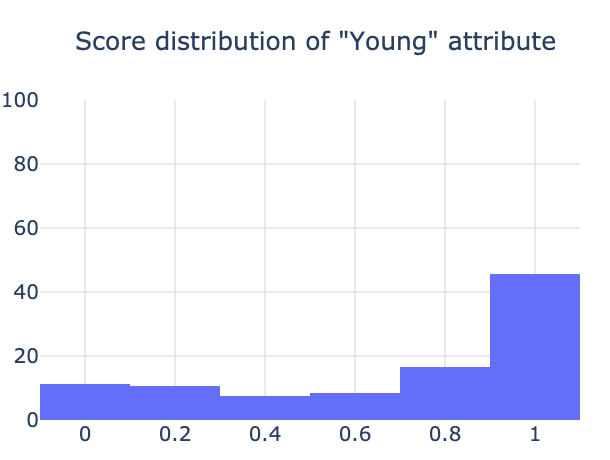}
}
\caption{Marginal probability distributions of `male`, `smiling`, `eyeglasses`, `young` attributes sampled from images generated by StyleGAN2 pre-trained on the FFHQ dataset.}
\label{fig:dist-marg}
\end{center}
\end{figure*}

\begin{figure*}
\begin{center}
\includegraphics[width=.33\textwidth]{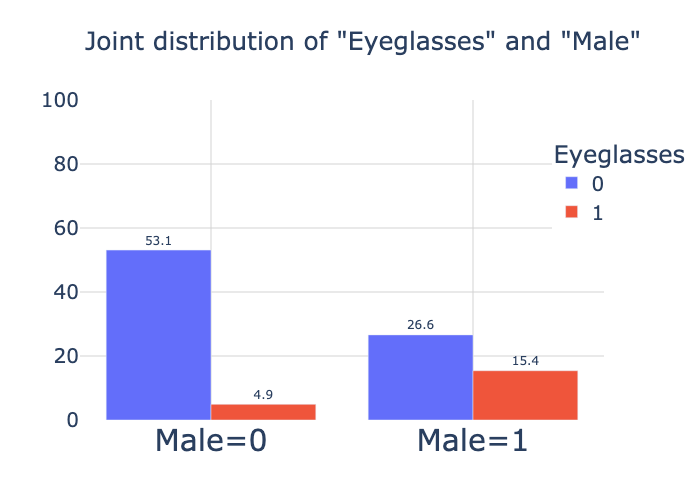}
\includegraphics[width=.33\textwidth]{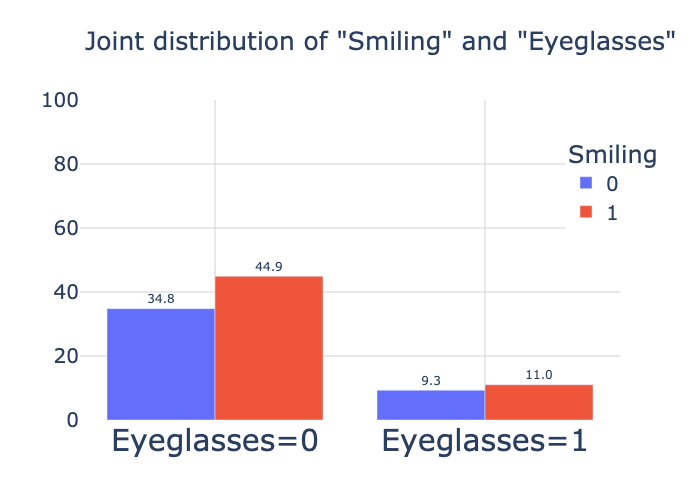}
\includegraphics[width=.33\textwidth]{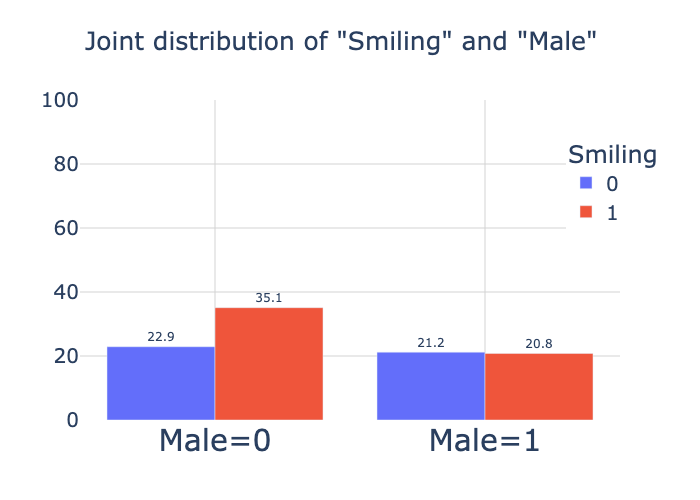}
\vskip -0.2in
\caption{Joint probability distributions of (`male`, `eyeglasses`), (`eyeglasses`, `smiling`), (`male`, `smiling`) attribute pairs sampled from images generated by StyleGAN2 pre-trained on the FFHQ dataset.}
\label{fig:dist-joint}
\end{center}
\vskip -0.1in
\end{figure*}

To understand how fair the StyleGAN2 model works on FFHQ, we randomly generated 1000 images. Then we used binary classifiers to label each image for the attributes \textit{gender, smiling, eyeglasses}, and \textit{young} for marginal and joint distributions (Table 3, Table 4). As can be seen, the StyleGAN2 model generates images that are slightly biased towards \textit{Male=False}, moderately biased towards \textit{Smiling=True} and strongly biased towards \textit{Young=True} and \textit{Eyeglasses=False} attributes.We also examine the joint distribution of attribute pairs such as \textit{gender + eyeglasses}, \textit{gender + smiling} and \textit{eyeglasses + smiling}. As can be seen, the joint probability distribution of the attributes can be extremely imbalanced even if the marginal probability distributions of the individual attributes are not, such as the ratio of \textit{women + eyeglasses} to \textit{men +  eyeglasses}. In Figure \ref{fig:dist-marg} and Figure \ref{fig:dist-joint}, respectively, we show the percentage of assigned binary labels for single and multiple attributes.

\section{Additional debiasing results}
\label{sec:appendix_add}
\begin{figure*}
\begin{center}
\includegraphics[width=1\textwidth]{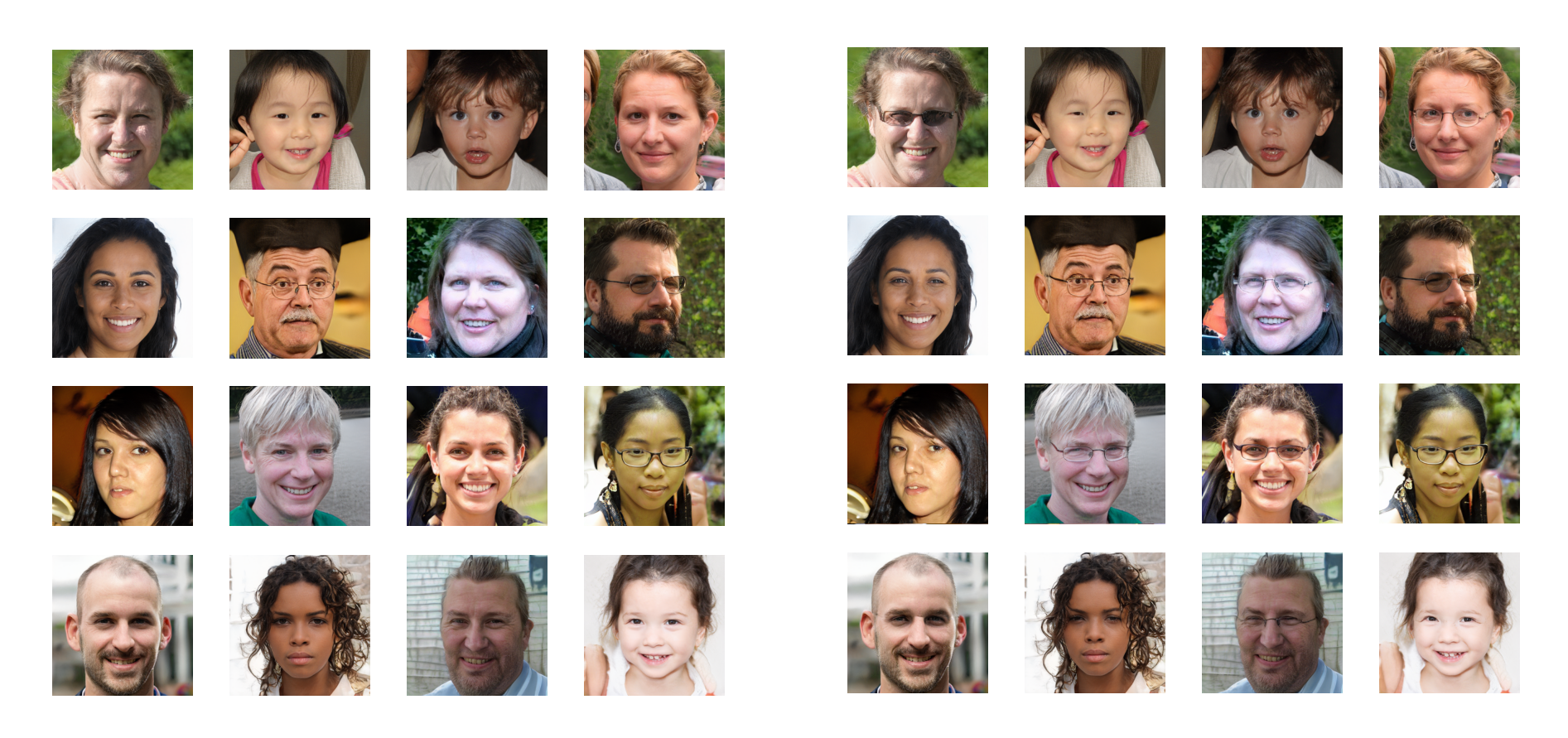}
\vskip -0.2in
\caption{A set of images generated with the same latent codes before and after debiasing the StyleGAN2 model with respect to the 'Eyeglasses' attribute on a single channel with our method.}
\label{fig:samelatent-eyeglasses}
\end{center}
\vskip -0.1in
\end{figure*}

\begin{figure*}
\begin{center}
\includegraphics[width=1\textwidth]{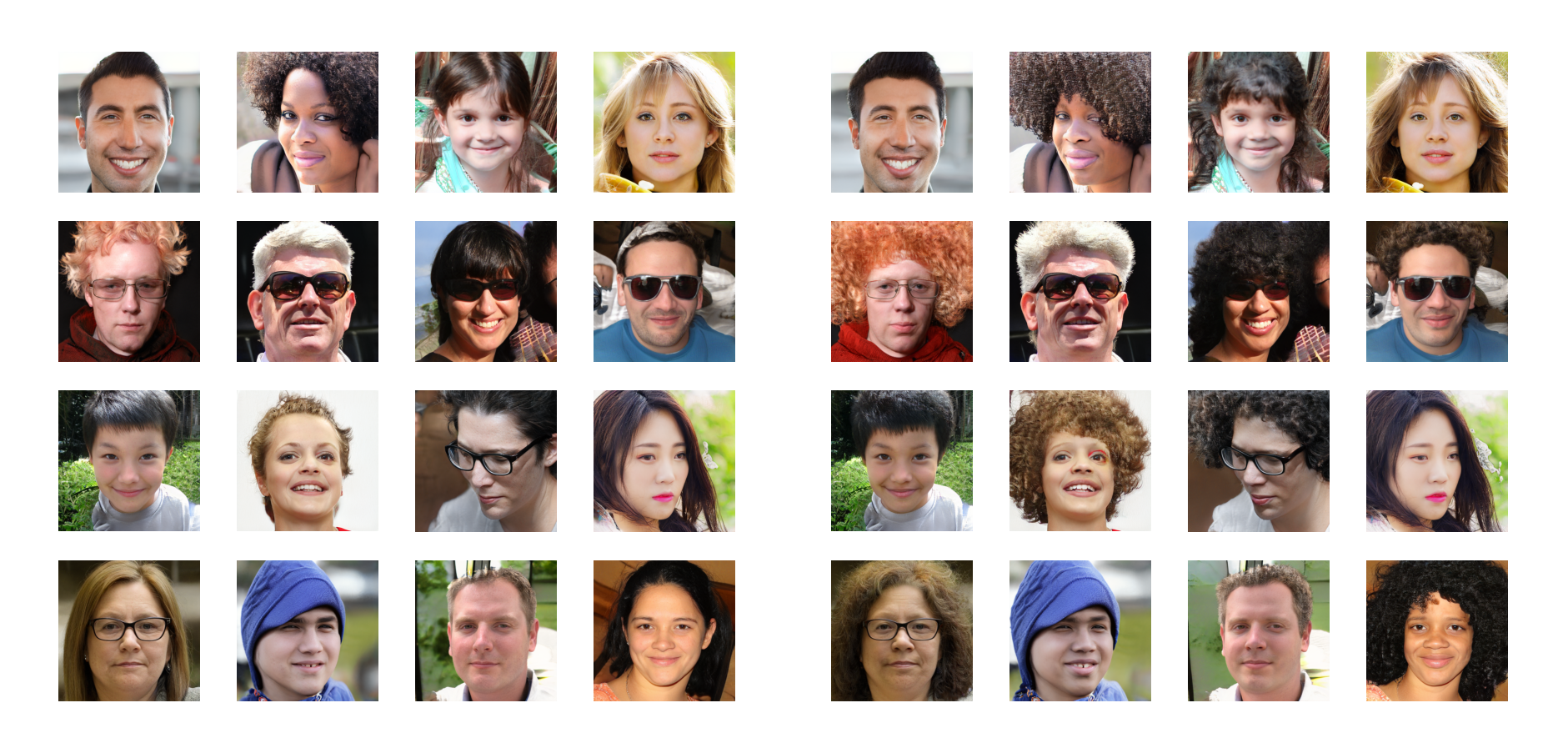}
\vskip -0.2in
\caption{A set of images generated with the same latent codes before and after debiasing the StyleGAN2 model with respect to the 'a person with afro hairstyle' text-based attribute with our method.}
\label{fig:samelatent-afrohair}
\end{center}
\vskip -0.1in
\end{figure*}

We also performed debiasing for  \textit{eyeglasses} (Figure \ref{fig:samelatent-eyeglasses}) and \textit{afro hair} attribute (Figure \ref{fig:samelatent-afrohair}) on the same latent codes showing the before/after of our debiasing method.

\begin{table*} [h]
\label{tab:dist-marg}
\caption{Marginal distributions of attributes measured on the FFHQ dataset and images generated by StyleGAN2 pretrained on the FFHQ dataset. }
\vspace{0.2em}
\begin{tabular}{p{0.23\columnwidth} | p{0.31\columnwidth} | p{0.31\columnwidth}}
\hline Attribute & FFHQ & StyleGAN2  \\
\hline Eyeglasses &  F=0.78, T=0.22 & F=0.80, T=0.20  \\
Young & F=0.28, T=0.72 & F=0.30, T=0.70 \\
Smiling & F=0.43, T=0.57 & F=0.44, T=0.56  \\
Male & F=0.58, T=0.42 & F=0.58, T=0.42  \\
\hline
\end{tabular}
\end{table*}

\begin{table*} [h]
\label{tab:dist-joint}
\caption{Joint distributions of attribute pairs measured on the FFHQ dataset and images generated by StyleGAN2 pretrained on the FFHQ dataset. }
\vspace{0.2em}
\begin{tabular}{p{0.23\columnwidth} | p{0.31\columnwidth} | p{0.31\columnwidth}}
\hline Attributes & FFHQ & StyleGAN2  \\
\hline Eyegl.-Smile & \shortstack{FF=0.34, FT=0.44  \\TF=0.09, TT=0.13} & \shortstack{FF=0.35, FT=0.45  \\TF=0.09, TT=0.11}   \\
\hline Smile-Male & \shortstack{FF=0.22, FT=0.36  \\TF=0.21, TT=0.21} & \shortstack{FF=0.23, FT=0.35  \\TF=0.21, TT=0.21}  \\
\hline Male-Eyegl. & \shortstack{FF=0.50, FT=0.08  \\TF=0.28, TT=0.14} & \shortstack{FF=0.53, FT=0.05  \\TF=0.27, TT=0.15}  \\
\hline
\end{tabular}
\end{table*}

\end{document}